\documentclass[10pt,twocolumn,letterpaper]{article}

\usepackage{cvpr}

\makeatletter
\@namedef{ver@everyshi.sty}{}
\makeatother
\usepackage{tikz}

\usepackage{times}
\usepackage{epsfig}
\usepackage{graphicx}
\usepackage{amsmath}
\usepackage{amssymb}
\usepackage{dsfont}
\usepackage{color}
\usepackage{comment}
\usepackage{subfig}
\usepackage{wrapfig}
\usepackage{nicefrac}
\usepackage{float}
\usepackage[normalem]{ulem}
\usepackage{animate}
\usepackage[export]{adjustbox}
\usepackage{enumitem}
\usepackage{multicol}
\usepackage{multirow}
\setlist{nosep} % For compact itemizations.

\usepackage{microtype}
\usepackage[pagebackref=true,breaklinks=true,letterpaper=true,colorlinks,bookmarks=false]{hyperref}
\usepackage{authblk}

\newcommand*\ccc[1]{\hspace{-2.5pt}{{ \tikz[baseline=(char.base)]{\node[shape=circle,draw,inner sep=1pt](char){#1};}}}\xspace}

\cvprfinalcopy

\definecolor{darkgreen}{RGB}{30,150,30}
\definecolor{darkblue}{RGB}{0,0,127}
\definecolor{darkyellow}{RGB}{171,133,0}
\definecolor{darkred}{RGB}{180,20,20}
\definecolor{darkmagenta}{RGB}{127,0,127}
\definecolor{darkcyan}{RGB}{0,127,127}

\hypersetup{
	linkcolor    = darkred, %Color of internal links
	citecolor   = darkgreen %Color of citations
}

\DeclareMathOperator*{\argmin}{arg\,min}

\begin{document}

%%%%%%%%% TITLE
\title{{Bi3D}: {S}tereo Depth Estimation via Binary Classifications}

\makeatletter
\renewcommand\Authfont{\fontsize{11.5}{14.4}\selectfont}
\renewcommand\AB@affilsepx{\qquad \protect\Affilfont}
\makeatother
\author[1,2]{Abhishek Badki}
\author[1]{Alejandro Troccoli}
\author[1]{Kihwan Kim}
\author[1]{Jan Kautz}
\author[2]{Pradeep Sen}
\author[1]{Orazio Gallo}
\affil[1]{NVIDIA}
\affil[2]{University of California, Santa Barbara}
\renewcommand*{\Authsep} { \ \ \ \ \  }%
\renewcommand*{\Authands}{ \ \ \ \ \  }%

\twocolumn[{%
\vspace{-7mm}
\renewcommand\twocolumn[1][]{#1}%
\maketitle
\begin{center}
    \centering
    \vspace*{-9mm}
    \captionsetup{type=figure}
    \includegraphics[width=\textwidth]{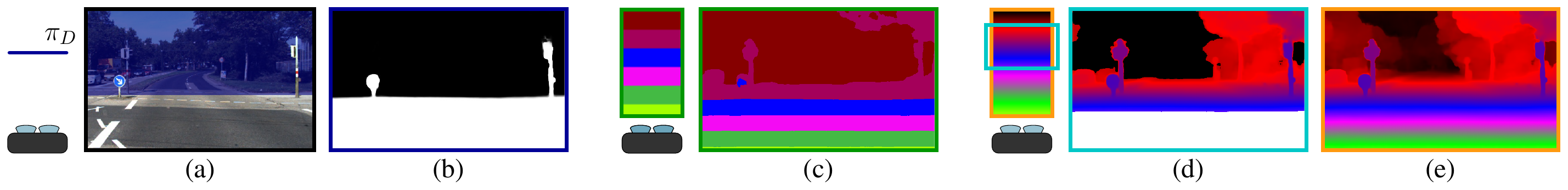}
    \captionof{figure}{Our stereo algorithm, named Bi3D, offers a trade-off between depth accuracy and latency. Given the plane at depth $D$ shown in the top view on the left, and overlaid on the scene in (a), our algorithm can classify objects as being closer (white) or farther (black) than $D$ in just a few milliseconds. Bi3D can estimate depth with arbitrary quantization, and complexity linear with the number of quantization levels, (c). It can also produce continuous depth within the same computational budget by focusing only a specific range, cyan region in the top view of (d), where pixels outside the range are still identified as closer (white) or farther (black). Finally, it can estimate the full depthmap, (e).
    \label{fig:teaser}}
\end{center}%
}]

\maketitle

%%%%%%%%% ABSTRACT
\begin{abstract}
%!TEX root = bi3d_cvpr20.tex

Stereo-based depth estimation is a cornerstone of computer vision, with state-of-the-art methods delivering accurate results in real time.
For several applications such as autonomous navigation, however, it may be useful to trade accuracy for lower latency.
We present Bi3D, a method that estimates depth via a series of binary classifications.
Rather than testing if objects are \emph{at} a particular depth $D$, as existing stereo methods do, it classifies them as being \emph{closer} or \emph{farther} than $D$.
This property offers a powerful mechanism to balance accuracy and latency.
Given a strict time budget, Bi3D can detect objects closer than a given distance in as little as a few milliseconds, or estimate depth with arbitrarily coarse quantization, with complexity linear with the number of quantization levels.
Bi3D can also use the allotted quantization levels to get continuous depth, but in a specific depth range.
For standard stereo (\ie, continuous depth on the whole range), our method is close to or on par with state-of-the-art, finely tuned stereo methods.{\let\thefootnote\relax\footnote{This work was done while A. Badki was interning at NVIDIA.}}
{\let\thefootnote\relax\footnote{Code available at \url{https://github.com/NVlabs/Bi3D}.}}
\end{abstract}

%%%%%%%%% BODY TEXT
\section{Introduction}\label{sec:intro}
%!TEX root = bi3d_cvpr20.tex

Stereo-based depth estimation is a core task in computer vision~\cite{scharstein2002taxonomy}.
State-of-the-art stereo algorithms estimate depth with good accuracy, while maintaining real-time execution~\cite{duggal2019deeppruner, khamis2018stereonet}.

Applications such as autonomous navigation, however, do not always require centimeter-accurate depth: detecting an obstacle within the braking distance of the ego vehicle (with the appropriate bound on the error) can trigger the appropriate response, even if the obstacle's depth is not \emph{exactly} known.
Moreover, the required accuracy and the range of interest varies with the task.
Highway driving, for instance, requires longer range but can deal with a more coarsely quantized depth than parallel parking.
Given a time budget for computing depth information, then, one can leverage this trade-off.

Unfortunately, existing methods do not offer the flexibility to adapt the depth quantization levels to determine if an object is within a certain distance, or simply to focus only on a particular range of the scene, without estimating the full depth first.
This is because, at their core, most existing algorithms compute depth by testing a number of candidate disparities, and by selecting the most likely under some cost function.
This results in two requirements on the choice of the candidate disparities for existing methods:
\begin{enumerate}
    \item they need to span a range covering all the objects in the scene, and 
    \item they cannot be arbitrarily coarse.
\end{enumerate}
If an object is outside the range spanned by the candidate disparities, existing methods still map it to the candidate disparity with the lowest cost, as shown in Figure~\ref{fig:results_selective}.
If the disparity candidates are too coarse, \ie, they are separated by too many disparity levels, the correct disparity may never be sampled, which, once again, results in the wrong depth estimation. 

\setlength{\fboxsep}{0pt}
\setlength{\fboxrule}{1pt}
\newcommand{\intuitionH}{.95in}
\begin{figure*}
\centering
\subfloat[]{\includegraphics[height=\intuitionH, trim={65 0 80 0}]{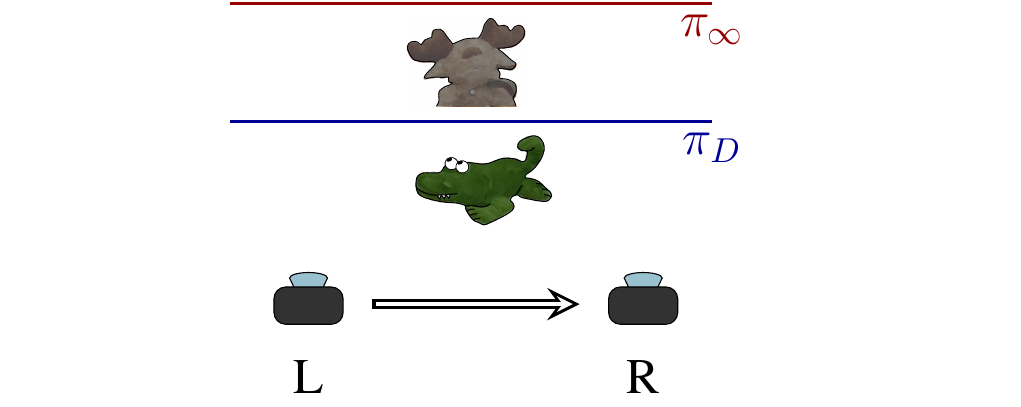}}~
\subfloat[L / R (Animated)]{\fcolorbox{black}{white}{\animategraphics[height=\intuitionH, poster=first]{2}{figures/intuition/input_}{01}{02}}}~
\subfloat[R]{\includegraphics[height=\intuitionH]{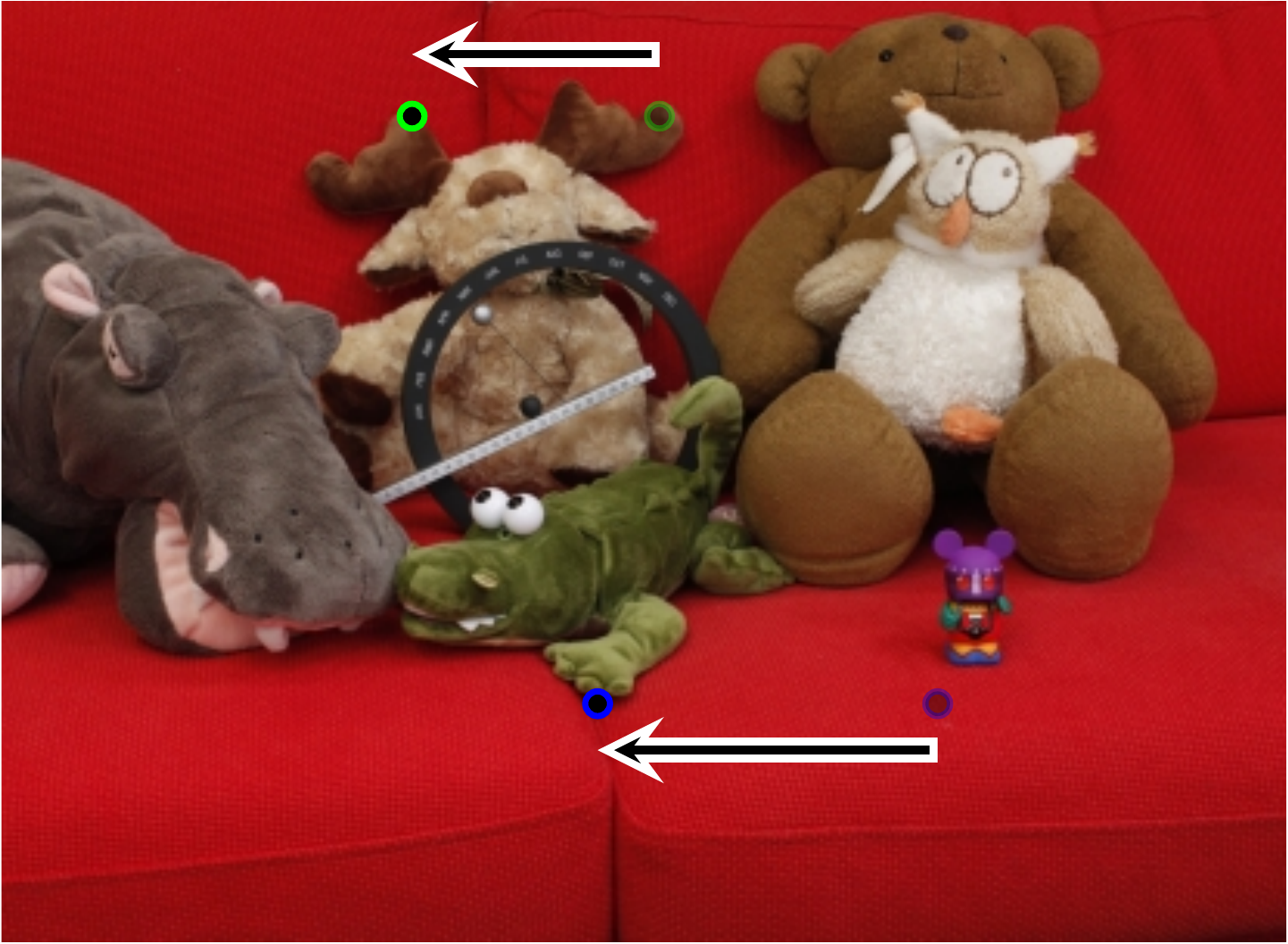}}~
\subfloat[L / Warped R (Animated)]{\fcolorbox{black}{white}{\animategraphics[height=\intuitionH, poster=first]{2}{figures/intuition/crop_}{01}{02}}}~
\subfloat[Warped R]{\includegraphics[height=\intuitionH]{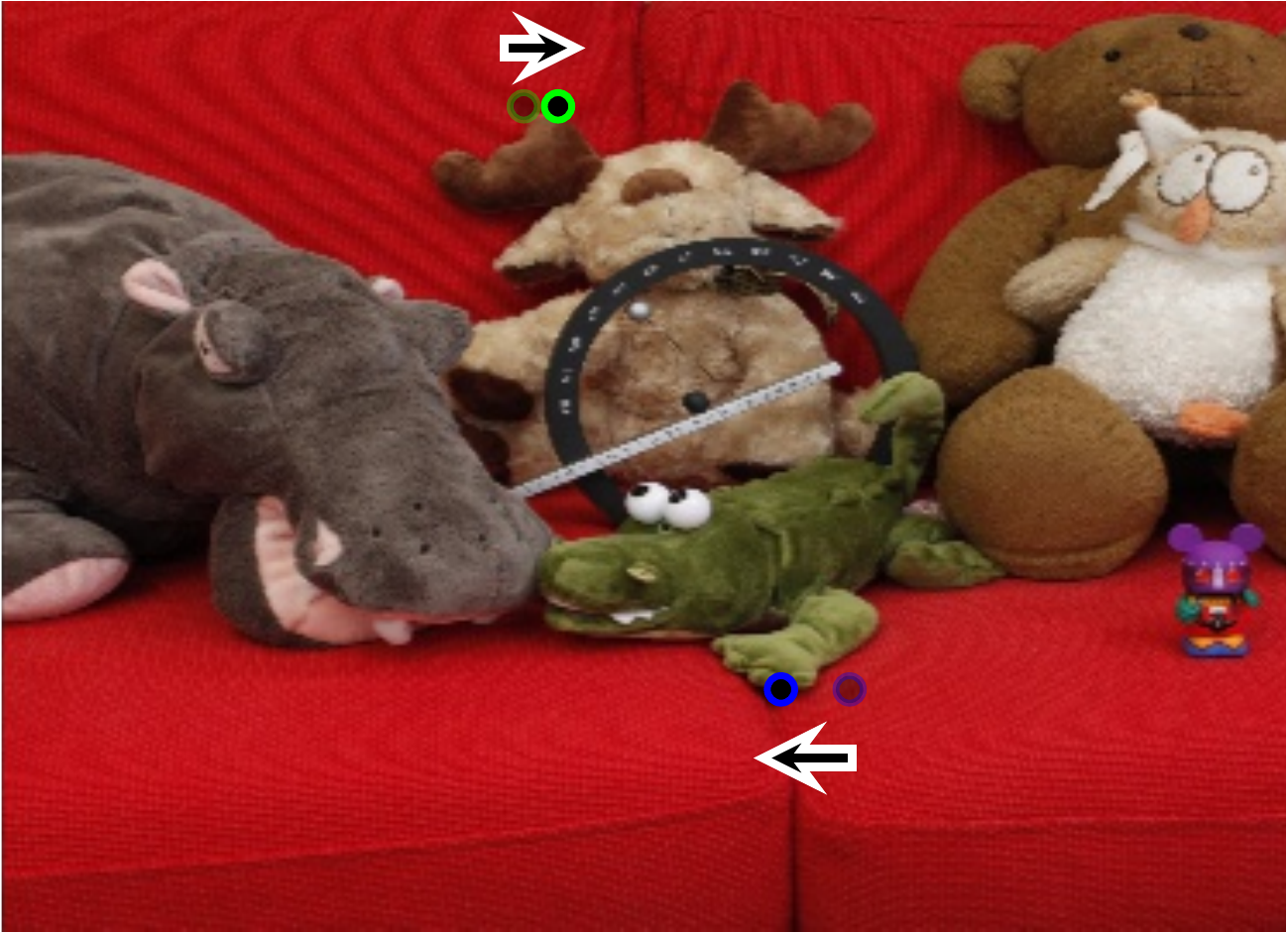}}
\vspace{-2mm}
\caption{Disparity, the apparent displacement of an object imaged by two cameras, is inversely proportional to the object's depth. Depth, then, can be estimated by regressing the magnitude of the disparity vectors. This is the operating principle of existing algorithms.  The direction of the disparity vectors, however, is the same for all the objects in the scene. After warping the right image to the left image via a plane in space $\pi_D$, the disparity of objects on opposite sides of the plane points to opposite directions. We propose to use this cue to estimate high-quality depth by classifying the direction of the disparity at multiple planes. Input images courtesy of Kim~\etal~\cite{kim2013scene}. \textbf{Figures (b) and (d) are animated. Please view in Adobe Reader and click on them to see the animation.}}\label{fig:intuition}
\vspace{-3mm}
\end{figure*}

We present a stereo method that allows us to take advantage of the trade-off between the depth quantization and the computational budget.
Our approach works by estimating the \emph{direction} of disparity relative to a given plane $\pi$, rather than regressing the absolute disparity, as existing methods do.
In other words, we learn to classify points in space as ``in front'' or ``behind'' $\pi$.
Plane $\pi$ can be regarded as a geo-fence in front of the stereo camera that can be used to detect objects closer than a safety distance.
By testing multiple such planes, we can estimate the depth at which a pixel switches from being ``in front'' to being ``behind,'' that is, the depth for that pixel.
Note that, because our method does not need to test the plane \emph{at or around} the actual depth of the pixel, the planes tested can be arbitrarily far from each other, allowing to control the depth's quantization.
For the same reason, our method offers valuable information even for objects that are outside of the tested range---whether they are in front or beyond the range.

To illustrate the core intuition of our approach, we consider the case of a camera pair, as in Figure~\ref{fig:intuition}(a).
Figures~\ref{fig:intuition}(b) and \ref{fig:intuition}(c) show that the magnitude of the disparity vector carries information about depth, with larger disparities indicating objects closer to the camera.
The disparity vector direction, however, does not change: all objects appear to ``move'' towards the left.

Figures~\ref{fig:intuition}(c) and \ref{fig:intuition}(d) show what happens when we warp the right image by the homography induced by $\pi_D$, the plane at distance $D$.
We note that now objects appear to be ``moving'' in different directions: the disparity vector direction for objects at, or in front of plane $\pi_D$ is the same as before, while for objects beyond plane $\pi_D$ it flips: it is now towards the right.
We leverage this observation and classify the direction of the disparity vector, rather than accurately regressing its magnitude.

If we repeat this task for several planes progressively closer to the camera, the depth of the plane at which the direction of the parallax vector associated with a 3D point flips yields the depth of the point.
A disparity vector that never flips indicates a 3D point that is in front or beyond all of the tested planes---it is outside of the search range.
Note, however, that the direction does tell us if the point is in front of or beyond the search range, which is valuable information.
Standard methods, in contrast, will generally assign each pixel a depth within the search range, even if the true depth is outside of this range.

In this paper we show that Bi3D, our stereo depth estimation framework, offers flexible control over the trade-off between latency and depth quantization:
\begin{itemize}
    \item Bi3D can classify objects as being closer or farther than a given distance in a few milliseconds. We call this \emph{binary depth estimation}, Figure~\ref{fig:teaser}(b).
    \item When a larger time budget is available, Bi3D can compute depth with varying quantization and execution time growing linearly with the number of levels. We refer to this as \emph{quantized depth}, Figure~\ref{fig:teaser}(c).
    \item Alternatively, it can estimate continuous depth in a range $[\pi_{D_1}, \pi_{D_2}]$ while identifying objects outside of this range as closer or farther than the extremes of the range. We refer to this as \emph{selective depth estimation}, Figure~\ref{fig:teaser}(d).
    \item Finally, Bi3D can estimate the full depth with quality comparable with the state-of-the-art.
\end{itemize}

\section{Related Work}\label{sec:related}
%!TEX root = bi3d_cvpr20.tex

Stereo correspondence has been one of most researched topics in computer vision for decades.
Scharstein and Szeliski present a good survey and provide a taxonomy that enables the comparison of stereo matching algorithms and their design~\cite{scharstein2002taxonomy}.
In their work, they classify and compare stereo methods based on how they compute their matching cost, aggregate the cost over a region, and optimize for disparity.
The matching cost measures the similarity of image patches.
Some metrics are built on the brightness constancy assumption, \eg, sum of squared differences or absolute differences~\cite{hirschmuller2007evaluation}.
Others establish similarity by comparing local descriptors~\cite{geiger2010efficient, zabih1994non}.
The matching costs can be compared locally, selecting the disparity that minimizes the cost regardless of the context; or more globally, using graph cuts~\cite{kolmogorov2001computing}, belief propagation~\cite{klaus2006segment}, or semi-global matching~\cite{hirschmuller2005accurate}.

A common limitation of most stereo algorithms is the requirement that disparities be enumerated.
This enumeration could take the form of a search range over the scanline on a pair of rectified stereo images, or the enumeration of all possible disparity matches in 3D space using a plane sweep algorithm~\cite{collins1996space}.
The result of computing the matching cost over a plane sweep volume is a cost volume in which each cell in the volume has a matching cost.
Cost volumes are amenable to filtering and regularization due to their discrete nature, which makes them powerful for stereo matching.

The advancement of neural networks for perception tasks also resulted in a new wave of stereo matching algorithms.
A deep neural network can be trained to compute the matching cost of two different patches, as done by Zbontar and LeCun~\cite{zbontar2015computing}.
Furthermore, a neural network can be trained to do disparity regression, as shown by Mayer~\etal~\cite{mayer2016large}.
But deep-learning methods can do more than matching and direct disparity regression.
Recent work on stereo matching trained on large datasets, such as the SceneFlow~\cite{mayer2016large} and KITTI~\cite{Menze2015kittidataset} datasets can compute features, their similarity cost, and can regularize the cost volume within a single end-to-end trained model.
For example, GC-Net regularizes a cost volume using 3D convolutions and does disparity regression using a differentiable soft argmin operation~\cite{kendall2017end}.
DPSNet leverages geometric constraints by  warping features into a cost volume using plane induced homographies, the same operation that is used to build plane sweep volumes~\cite{im2019dpsnet}.
Aggregation of context information is important to handle smooth regions and repeated patterns; PSM-Net~\cite{chang2018pyramid} can take advantage of larger context through spatial pyramid pooling.
GA-Net~\cite{zhang2019ga} improves cost aggregation by providing semi-global and local cost aggregation layers.
The downside of large architectures is their computational cost, which makes them unsuitable for real-time applications.
To overcome this, architectures like DeepPruner~\cite{duggal2019deeppruner} reduce the cost of volume matching by pruning the search space with a differentiable PatchMatch layer.

To summarize, most of the existing work in stereo correspondence is based on the computation of a discrete cost volume or fixing a disparity search range.
The exception is DispNet~\cite{mayer2016large}, but its performance in the KITTI stereo benchmark has been surpassed by others.
Our key contribution is the framing of depth estimation as a collection of binary classification tasks. 
Each of these tasks provides useful depth information about the scene by estimating the upper or lower bound on the disparity value at each pixel.
As a result, our proposed method can be more selective or adaptive according to the task and the scene.
We can also do accurate disparity regression by repeating the binary classifications.

\section{Method}\label{sec:method}
%!TEX root = bi3d_cvpr20.tex

Given the left and right image of a stereo pair, which we indicate in the following as reference ($L$) and source ($R$), respectively, we can build a plane sweep volume (PSV) by selecting a range of disparities $\{d_i\}_{i=0:N}$.
Each plane of the PSV with reference to the left image can be computed as
\begin{equation}\label{eq:PSV}
\text{PSV}(x,y,d_i) = \mathcal{W}(R(x,y); H_{\pi_{d_i}}),
\end{equation}
where, with slight abuse of notation, $\mathcal{W}(\,\cdot\,; H)$ is a warping operator based on homography $H$. $H_{\pi_{d_i}}$ is the homography induced by the plane at the depth corresponding to disparity $d_i$.
(In the following we refer to $\pi_{d_i}$ as the plane at disparity $d_i$.)

Given a matching cost $C$, then, existing algorithms estimate the disparity for a pixel as
\begin{equation}\label{eq:cost_based_disparity}
\hat{d}(x,y) = \argmin_{d_i} C(\mathcal{N}_{L}(x,y), \mathcal{N}_\text{PSV}(x,y,d_i)),
\end{equation}

where $\mathcal{N}$ is a neighborhood of the pixel. 
The choice of the cost $C$ varies with the algorithm.
It can be a simple normalized-cross correlation of the grayscale patches centered at $(x,y)$, or it could be the output of a neural network and, thus, computed on learned features instead~\cite{kendall2017end}.
Regardless of the choice,
\begin{equation}\label{eq:predefined_disp}
\hat{d}(x,y) \in [d_0, d_N].
\end{equation}

Equations~\ref{eq:cost_based_disparity} and~\ref{eq:predefined_disp} imply that we need to evaluate all the candidate disparities before selecting the best match and an object whose disparity is outside of the range $D = [d_0, d_N]$ is still mapped to the interval $D$.
We argue that this is a major limitation and show how it can be lifted.

\subsection{Depth via Binary Classifications}\label{sec:method_core_idea}
\begin{figure}
\centering
\input{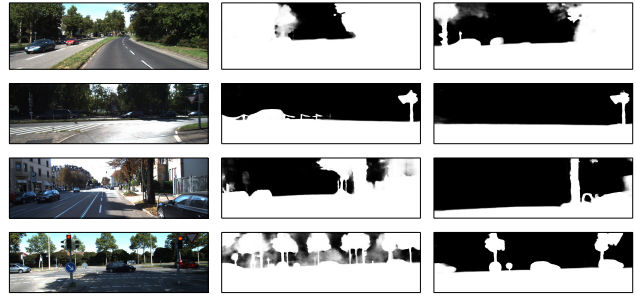}
\caption{Given a stereo pair and $\pi_{d_i}$, the plane corresponding to disparity ${d_i}$, our method can estimate whether an object is closer or farther than $\pi_{d_i}$. Each row shows the confidence maps for the scene on the left for two different disparity levels (white means ``in front'').}\label{fig:segmentation}
\vspace{-1mm}
\end{figure}

Instead of estimating $\hat{d}(x,y)$ directly, we observe that the \emph{direction} of the disparity vector itself carries valuable information.
After warping image $R$ with Equation~\ref{eq:PSV}, in fact, the disparity direction flips depending on whether the object is in front of $\pi_{d_i}$ or behind it.
The animation in Figure~\ref{fig:intuition} shows the stereo pair before and after warping.

This suggests that we can train a binary classifier to take two images, $L(x,y)$ and $\text{PSV}(x,y,d)|_{d=d_i}$, and predict the parts of the scene that are behind (or in front of) $\pi_{d_i}$.
To do this, we train a standard neural network using a binary cross-entropy loss (details on the architecture are offered in Section~\ref{sec:implementation}).
At convergence, the classifier's output can be remapped to $[0,1]$ yielding
\begin{equation}\label{eq:confidence}
\mathcal{C}_{d_i}(x,y) = \sigma(o(x,y)),
\end{equation}
where $o$ is the output of the network, and $\sigma(\cdot)$ is a sigmoid function.
With a slight abuse of notation, we refer to $\mathcal{C}_{d_i}$ as \emph{confidence}.
When $\mathcal{C}_{d_i}$ is close to $1$ or $0$, the network is confident that the object is in front or behind the plane, respectively, while values close to $0.5$ indicate that the network is less confident.
We can then classify the pixel by thresholding $\mathcal{C}$ at $0.5$ (see Section~\ref{sec:implementation} for more details).
Figure~\ref{fig:segmentation} shows examples of segmentation masks that this approach produces for a few images from the KITTI dataset~\cite{Menze2015kittidataset}.
We refer to this operation as \emph{binary depth} estimation.
Binary depth, though based on a single disparity, already provides useful information about the scene: whether an object is within a certain distance from the camera or not---a form of geo-fence in front of the stereo camera. 

Existing approaches cannot infer binary depth without computing the full depth first.
This is because they have to test a set of disparities to select the most likely.

\begin{figure}
\vspace{-1.5mm}
\centering
\includegraphics[width=\columnwidth]{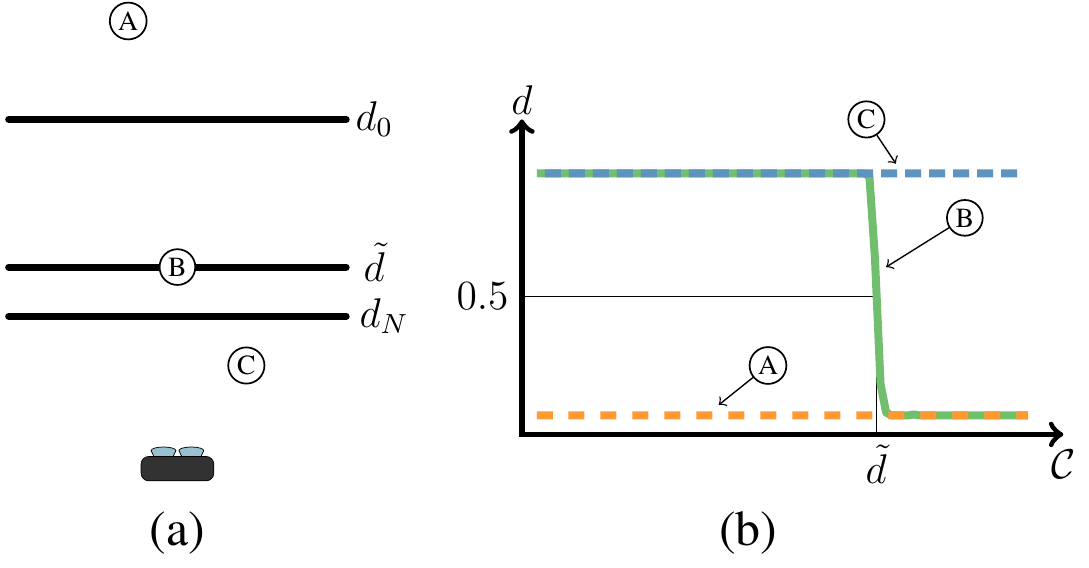}
\caption{The plots on the right are confidence values predicted by our network for objects in different positions relative to a disparity range of interest, $[d_0,d_N]$, as shown on the left. For objects within the range, the confidence will pass through the $0.5$ confidence level at the object's true disparity. For objects in front or beyond the range the confidence does not cross $0.5$ and stays around $1$ or $0$, respectively.
}\label{fig:plots}
\vspace{-1mm}
\end{figure}

We can repeat this classification for a set of disparity planes $\{d_i\}_{i=0:N}$ and concatenate the results for different disparities in a single volume $\mathcal{C}(x,y,d)|_{d=d_i}=\mathcal{C}_{d_i}(x,y)$, which we refer to as confidence volume.
Figure~\ref{fig:plots}(b) shows $\mathcal{C}(x,y,d)|_{(x,y)=(x_0,y_0)}$ (confidence across all the disparity planes for a particular pixel) for objects at different positions with respect to the range.
Assume that an object lies within the range of disparities $[d_0, d_N]$, like object \ccc{B}.
For disparity planes far behind the object, the classifier is likely to be confident in predicting that the object \ccc{B} is in front (\ie, $\mathcal{C}=1$).
Similarly, for disparity planes much closer than the object, we expect it to be confident in predicting the opposite (\ie, $\mathcal{C}=0$).
For disparity planes that are closer to the object, however, the prediction is less confident.
This is understandable: the closer the plane is to the object's depth, the smaller the magnitude of the disparity vector, making the direction difficult to classify.
In principle, at the correct disparity the classifier should be unsure and predict $0.5$.

However, because of unavoidable classification noise, simply taking the first disparity at which the curve crosses $0.5$ as our estimate could result in large errors.

To find the desired disparity, we could robustly fit a function and solve for the disparity analytically, or even train a zero-crossing estimation network.
However, we find the area under the curve (AUC) to be a simpler alternative that works well in practice:
\begin{equation}\label{eq:auc}
\hat{d}(x,y) = \sum_{d_i} \mathcal{C}(x,y,d_i)\cdot(d_i-d_{i-1}).
\end{equation}

To understand the intuition behind Equation~\ref{eq:auc}, consider the case in which the network is very confident for a particular pixel $(x_0,y_0)$:
\begin{equation}\label{eq:perfect_c}
%\vspace{-1mm}
\mathcal{C}(x_0,y_0,d) = 
\begin{cases}
1 &\mbox{for } d< \tilde{d}\\ 
0 &\mbox{for } d\geq \tilde{d}
\end{cases},
\end{equation}
where $\tilde{d}$ is the ground-truth disparity. 
Equation~\ref{eq:auc} estimates a disparity that matches $\tilde{d}$ up to a quantization error.
As shown in Figure~\ref{fig:plots}, $\mathcal{C}$ is roughly linear in the region around the correct disparity.
We show in the supplementary video that, under this condition, Equation~\ref{eq:auc} holds even for transitions that extend over a larger number of disparities.

Equation~\ref{eq:auc} has desirable properties.
First, it exhibits some tolerance to wrong confidence values.
Consider the case in which a few consecutive planes of the confidence volume of a pixel are classified confidently but incorrectly (\ie, $0$ instead of $1$ or vice versa).
An approach looking for the crossing of the $0.5$ value, may interpret the transition as the desired disparity, potentially yielding a completely wrong result.
The prediction in Equation~\ref{eq:auc} would just be shifted by a few disparity levels.
Second, the estimate of this operation need not be one of the tested disparities: it can be a continuous value.
This allows us to estimate sub-pixel accurate disparity maps when we use Equation~\ref{eq:auc} as a loss function during training. We explain this in Section~\ref{sec:implementation}.

Figure~\ref{fig:full_range_results} shows a comparison of the depth estimation results obtained with this method and other state-of-the-art stereo estimation approaches.
Note that the visual quality of our results is on par with GwcNet~\cite{Guo_2019_GwcNet} and GA-Net~\cite{zhang2019ga}.

\subsection{Coarsely Quantized Depth Estimation}\label{sec:quantized}
For some use-cases binary depth may not be enough, but full, continuous depth may be unnecessary.
Consider the case of highway driving, for instance: knowing that an obstacle is within, say, $1$ meter of the braking distance as quickly as possible may be preferable over knowing its exact distance with a higher latency.
Existing methods estimate depth by looking at all the candidate depths and use function such as softargmax to achieve sub-pixel disparity.
Therefore, we cannot change the coarsity at inference time or estimate depth with arbitrary quantization.
Given a fronto-parallel plane $\pi_{d_i}$, which splits the range into two segments, the true confidence $\mathcal{\widetilde{C}}_{d_i}(x,y)$ (that Bi3D estimates) is related to a cumulative distribution function, CDF:
\begin{equation}\label{eq:cdf}
    p(d(x,y) \leq d_i) = 1 - p(d(x,y) > d_i) = 1 - \mathcal{\widetilde{C}}_{d_i}(x,y),
\end{equation}
where $d(x,y)$ is the disparity of the pixel.
Note that this is a proper CDF, since $\mathcal{\widetilde{C}}_{d_i=0} = 1$ (everything is in front of the zero-disparity plane, \ie, plane at infinity) and $\mathcal{\tilde{C}}_{d_i=\infty} = 0$ (nothing is in front of the plane at depth zero).
Given a farther plane $\pi_{d_j}$, where $d_j<d_i$, then, we can write
\begin{equation}\label{eq:prob}
    p(d_j < d(x,y) \leq d_i) = \mathcal{\widetilde{C}}_{d_j}(x,y) - \mathcal{\widetilde{C}}_{d_i}(x,y).
\end{equation}
Equation~\ref{eq:prob} allows to estimate depth with arbitrary quantization.
To get $N+1$ quantization levels, we can use $N$ planes, compute the probability for each quantization bin using Equation~\ref{eq:prob} and estimate the pixel's disparity as the center of the bin $(d_j, d_i)$ that has the highest probability.
This amounts to treating quantized depth estimation as a hard-segmentation problem.
If we treat it as a soft-segmentation problem and assume uniform quantization bins, using this CDF-based approach naturally simplifies to the AUC method described in Section~\ref{sec:method_core_idea}.

The computational complexity is linear with the number of planes.
As shown in Table~\ref{tab:quantization_iou}, our method produces results that are on par with state-of-the-art methods at a fraction of the time, see Section~\ref{sec:results}. 

\begin{figure}
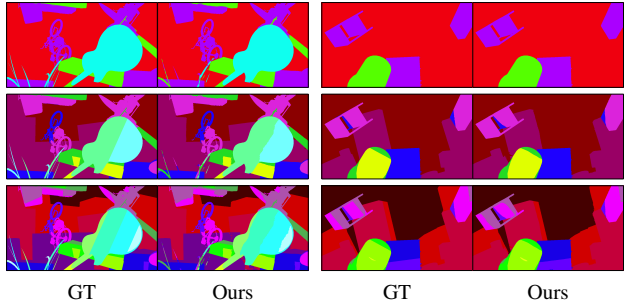

    \include{figures/quantized/quant_fig}
    \caption{Our method can estimate depth with arbitrarily coarse quantization. Rows one through three show the resulting depth maps for $4$, $8$, and $16$ levels, respectively. Note that even at $4$ levels one can get a basic understanding of the scene, but with much lower latency, see Table~\ref{tab:quantization_iou}.}\label{fig:coarse_depth}
    \vspace{-1mm}
\end{figure}

\begin{figure*}
    \centering
    \vspace{-2.5mm}
    \includegraphics[width=1.005\textwidth, trim={20 270 610 0}, clip]{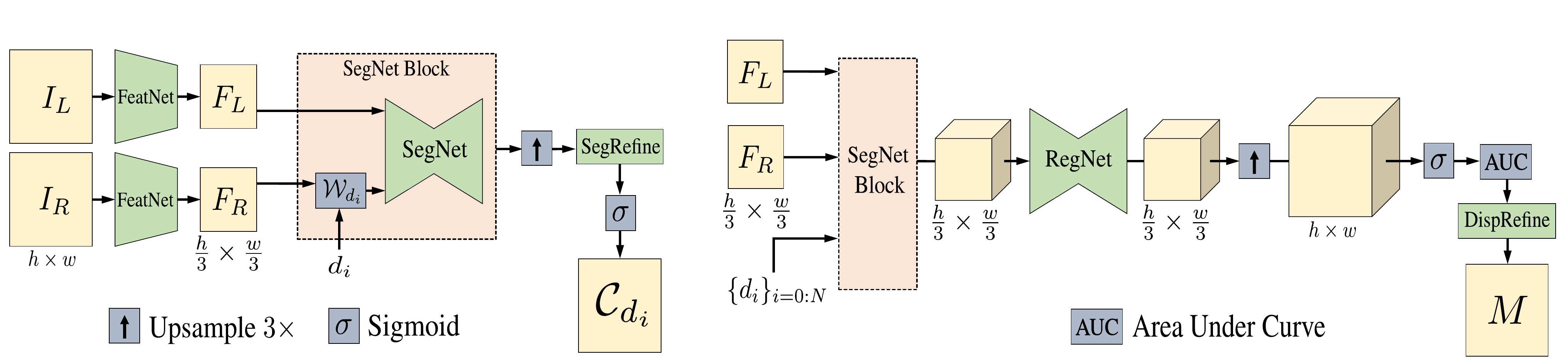}
    \vspace{-5mm}
    \caption{Bi3DNet, our core network, takes the stereo pair and a disparity $d_i$ and produces a confidence map (Equation~\ref{eq:confidence}), which can be thresholded to yield the binary segmentation (left). To estimate depth on $N+1$ quantization levels (Section~\ref{sec:quantized}) we run this network $N$ times and maximize the probability in Equation~\ref{eq:prob}. To estimate continuous depth, whether full or selective, we run the SegNet block of Bi3DNet for each disparity level and work directly on the confidence volume (right). }
    \label{fig:diagram}
    \vspace{-1mm}
\end{figure*}

\subsection{Selective Depth Estimation}\label{sec:selective}
\begin{figure*}
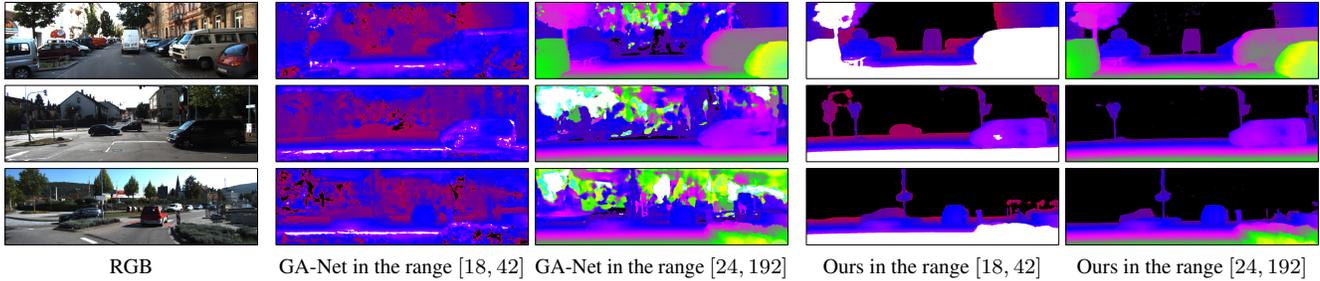

\vspace{-4mm}
\include{figures/selective_range/selective}
\caption{Example of selective depth estimation for three different scenes and with two different disparity ranges, indicated in the labels. Our method predicts the correct depth in the range of interest. White and black pixels are those detected as being in front or behind the selected range, respectively.}\label{fig:results_selective}
\vspace{-3mm}
\end{figure*}

So far, we have assumed that the range $[d_0, d_N]$ covers all the disparities in the scene.
Generally, this puts $d_0$ at $0$ (plane at infinity) and $d_N$ at the the maximum disparity expected in the scene.
Since this information is not available a priori, however, existing methods resort to using $192$ disparity levels, with each disparity level being $1$ pixel wide, and with $d_0=0$.
Now consider an object outside of the range in Figure~\ref{fig:plots}(a), such as \ccc{C}.
In this case, because they look for a minimum cost, existing methods are forced to map the object to a (wrong) depth within the range, see Figure~\ref{fig:results_selective}.
Even if they had a strategy to detect out-of-range objects, \eg, threshold the cost, the best they could do would be to acknowledge that no information about the depth of the object is known.

Our approach, on the other hand, deals with out-of-range objects seamlessly.
Because the direction of the disparity vector associated with an object such as \ccc{C} never changes, the confidence value $\mathcal{C}$ stays at $1$ throughout the range.
Therefore, in addition to knowing that \ccc{C} is outside of the range we are testing, we also know that it is in front of the closest plane.
If we move the farthest plane to a non-zero disparity, the same considerations apply to objects that may now be beyond the range, such as \ccc{A}. 
Thanks to this ability, our method is robust to incorrect selections of the range, unlike conventional methods.

However, we can leverage this property further.
Given a budget of disparity planes that we can test, rather than distributing them uniformly over the whole range, we can allocate them to a specific range.
We refer to this as \emph{selective depth}.
Selective depth allows us to get high quality, continuous depth estimation in a region of interest within the same computational budget of quantized depth.
Objects outside of this range are classified as either in front or behind the working volume. 
Given a disparity range $[d_\text{min}, d_\text{max}]$, then, we can apply the exact same method described in Section~\ref{sec:method_core_idea}.

Figure~\ref{fig:results_selective} shows selective depth results.
Note that our results look like regular depthmaps that are truncated outside of the selected range.
In comparison, GA-Net~\cite{zhang2019ga} struggles to map pixels outside the range to the correct values, which affects the confidence for objects that lie within the selected range as well.

\subsubsection{Adaptive Depth Estimation}\label{sec:adaptive}
\begin{figure*}
    \captionsetup[subfigure]{labelformat=empty}
    \centering
    \vspace{-5mm}
    \subfloat[Time $t_0$]{\includegraphics[width=.32\textwidth]{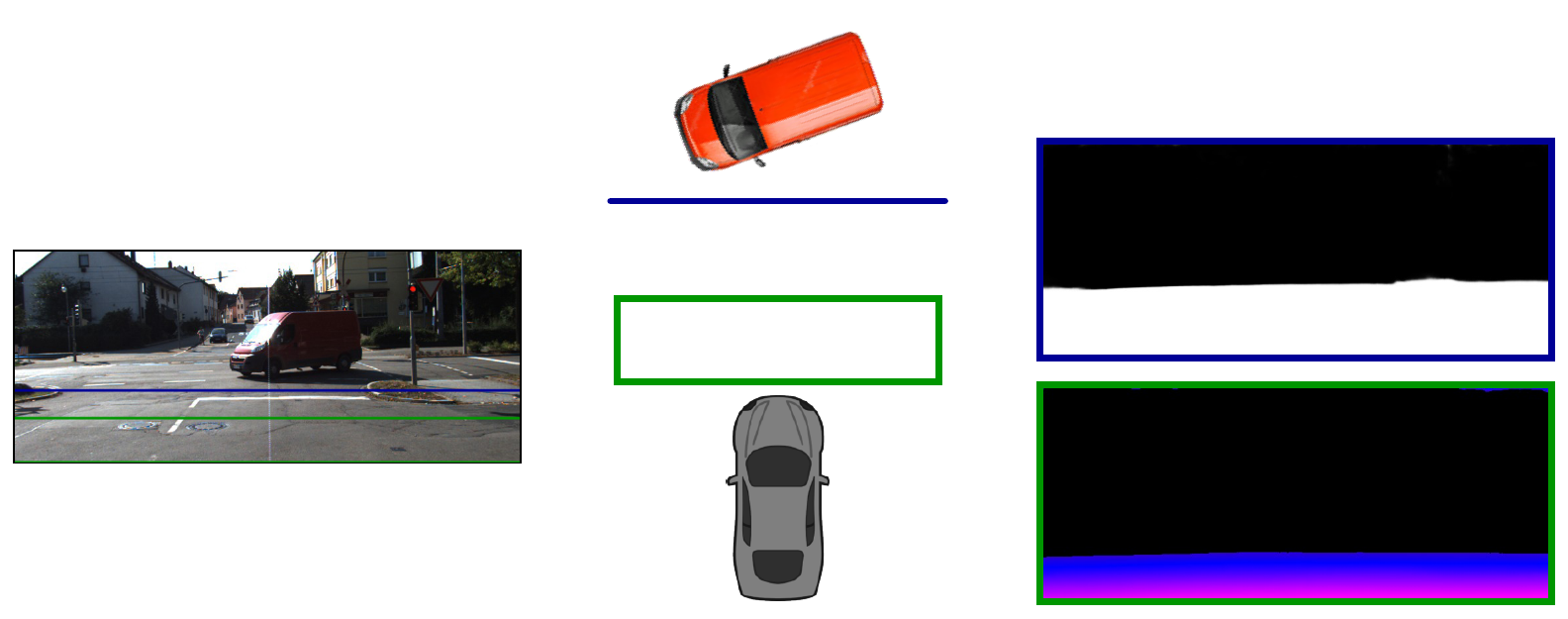}}~~~
    \subfloat[Time $t_1$]{\includegraphics[width=.32\textwidth]{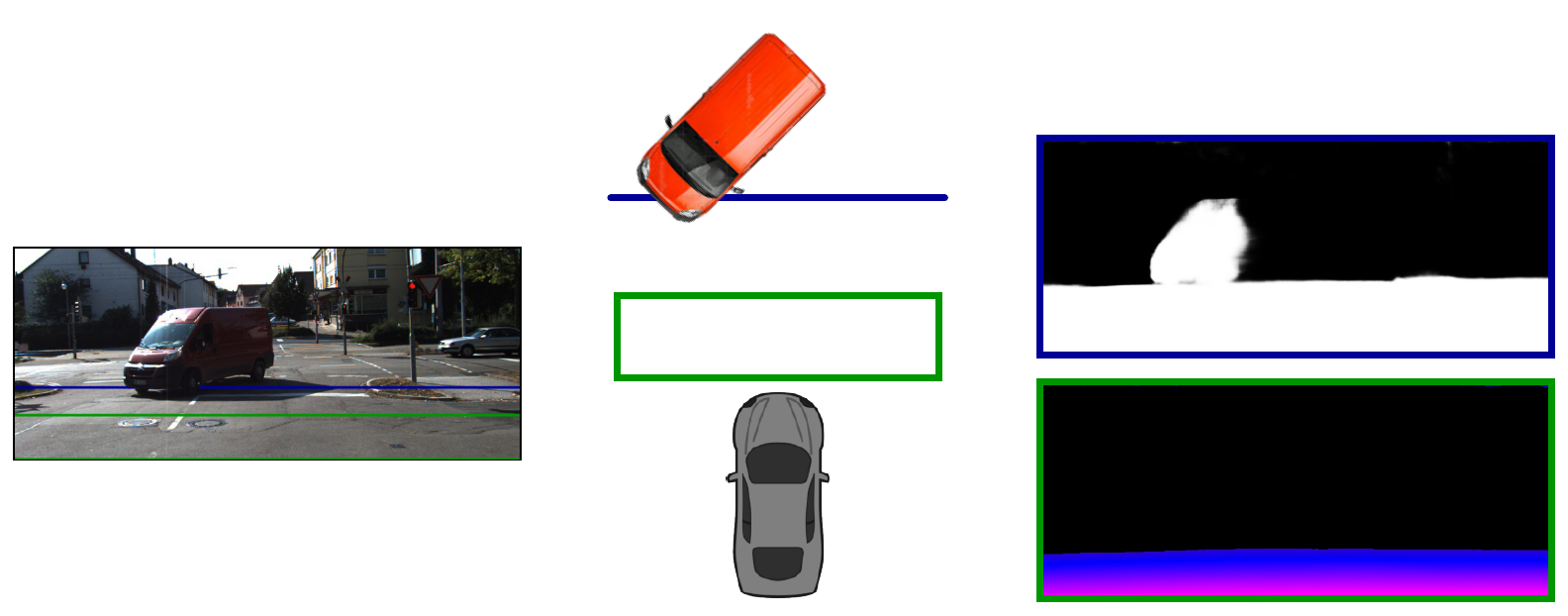}}~~~
    \subfloat[Time $t_2$]{\includegraphics[width=.32\textwidth]{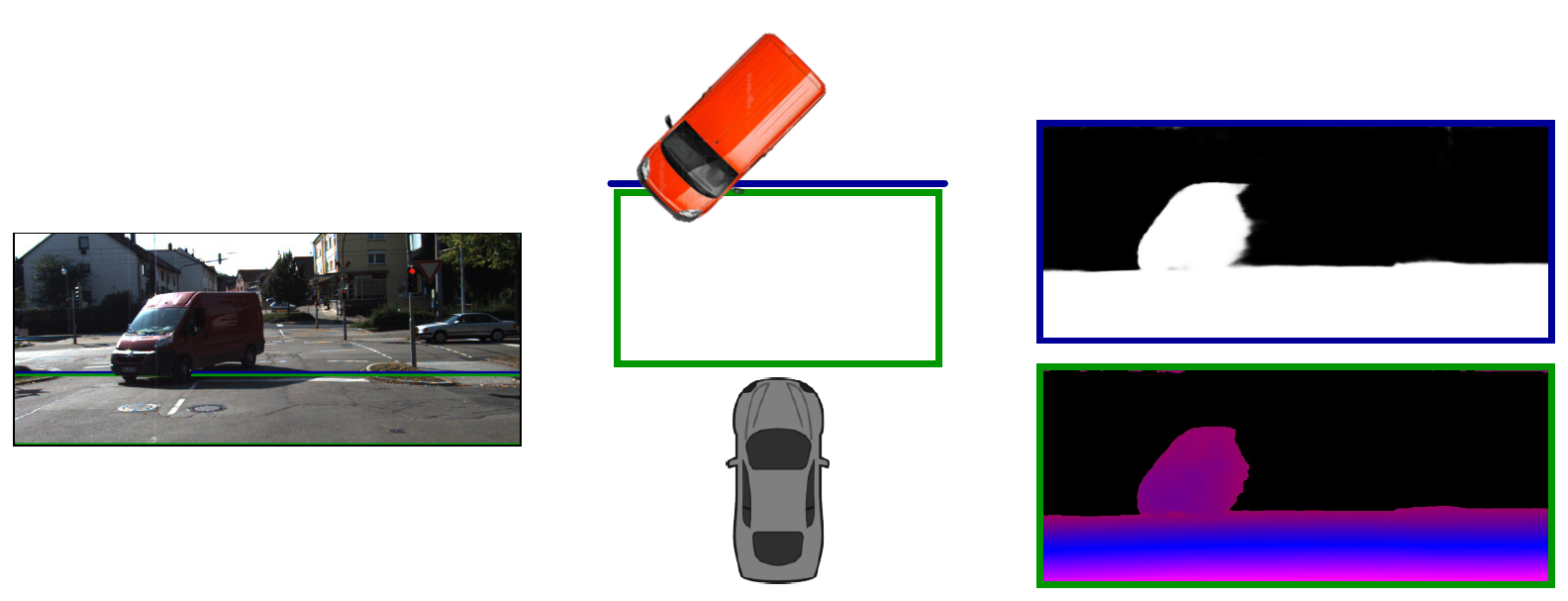}}
    \caption{By combining selective and binary depth estimation, we can implement a simple adaptive depth estimation strategy, here shown for an automotive application. We perform selective depth estimation for a close range (in green in the top view) to best use a given budget of disparity planes. We also monitor farther ranges with binary depth (blue plane). When an object crosses the far plane (see Time $t_1$), the selective range can be extended to estimate the distance to the van (see Time $t_2$).}\label{fig:adaptive_depth}
\end{figure*}

The advantages of these techniques can also be combined to minimize latency while maximizing the information about the scene. 
Consider the case of perception for autonomous navigation:
selective depth can be used for the region directly in front of the ego vehicle, yielding high depth quality in the region that affects the most immediate decisions.
However, information about farther objects cannot be completely dismissed.
Rather than using additional sensors, such as radar, farther objects can be monitored with binary depth, which would act as a geo-fence that moves with the ego vehicle.
This situation is depicted in Figure~\ref{fig:adaptive_depth} at time $t_0$, where the top view indicates the selective range in green, and the binary depth plane in blue. 
At time $t_1$, when the red van crosses the blue plane, the range of the selective depth can be extended yielding information about the van approaching.
This application is best show-cased with a video, which we provide in the supplementary video.

\begin{figure*}
\input{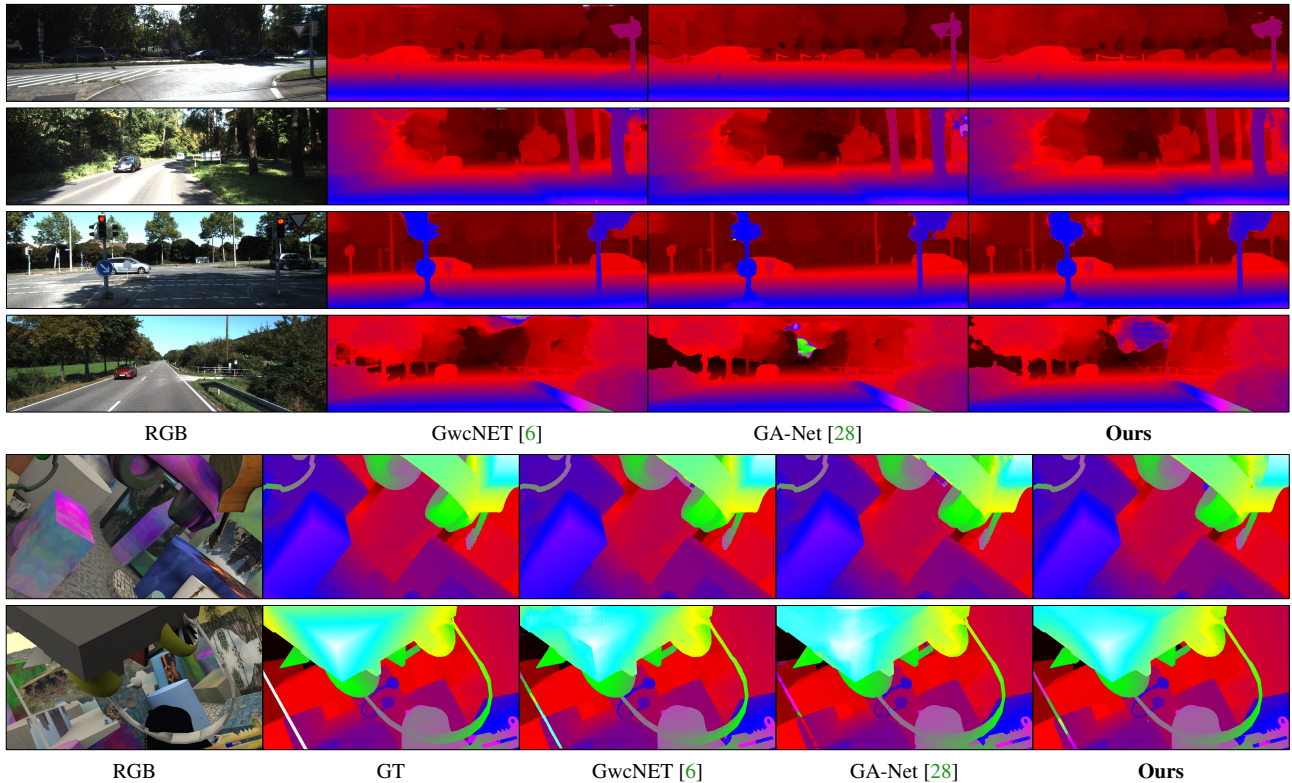}
\caption{Results for standard stereo for both KITTI~\cite{Menze2015kittidataset} and the Flying Things~\cite{mayer2016large} datasets. While our goal is an algorithm that allows flexible selection of the range, even on traditional stereo (\ie, looking for correspondences across the whole range) our method produces results visually comparable to the state-of-the-art.}\label{fig:full_range_results}
\end{figure*}

\begin{table*}
    \centering
    \scalebox{0.72}{
    \begin{tabular}{ccccccccccc}
    \hline
    GC-Net~\cite{kendall2017end} & DipsNetC~\cite{mayer2016large} &  CRL~\cite{Pang2017} & PDS-Net~\cite{Tulyakov2018} & PSM-Net~\cite{chang2018pyramid} & DeepPruner~\cite{duggal2019deeppruner} & GA-Net-15~\cite{zhang2019ga} & CSPN~\cite{cheng2018depth} & MCUA~\cite{nie2019multi} & GwcNet~\cite{Guo_2019_GwcNet}& \textbf{Ours} \\ 
    \hline
    2.51 & 1.68 & 1.32 & 1.12 & 1.09 & 0.86 & 0.84 & 0.78 & 0.56 & 0.76 & \textbf{0.73} 
    \end{tabular}
    }
    \caption{EPE values on the Scene Flow dataset for several state-of-the-art methods. Our method has the second best score.}
    \label{tab:resultssf}
\end{table*}

\section{Implementation}\label{sec:implementation}
%!TEX root = bi3d_cvpr20.tex

In this section we provide high-level details for our implementation.
More details are in the Supplementary.

The core of our method is a network that takes in a stereo pair and a disparity level $d_i$, and produces a binary segmentation with respect to $\pi_{d_i}$.
We call this network Bi3DNet, see Figure~\ref{fig:diagram} (left).
The first module, FeatNet, which extracts features from the stereo images, is a simplified version of the feature extractor of PSM-Net~\cite{chang2018pyramid}.
The output is a $32$-channel feature map at a third of the resolution of the original image. 
FeatNet needs to run only once, regardless of the number of disparity planes we seek to test.
The left-image features and the warped right-image features are then fed to SegNet, a standard $2$D encoder-decoder network with skip connections.
Our SegNet architecture downsamples the input $5$ times. 
At each scale in the encoder, we have a conv layer with stride of $2$ followed by a conv layer with stride of $1$. 
The decoder follows the same approach, where at each scale we have a deconv layer with a stride of $2$ followed by a conv layer with a stride of $1$.
A final conv layer estimates the output of SegNet, which we bilinearly upsample to the original resolution.
We then refine it with a convolutional module (SegRefine), which also takes the input left image for guidance (not shown in Figure~\ref{fig:diagram}).

Estimating continuous or quantized depth, then, simply requires to run Bi3DNet multiple times.
For quantized depth we use Bi3DNet directly, stack the $\mathcal{C}_{d_i}$'s, and maximize the probability in Equation~\ref{eq:prob}.
Note that, in order for the network to be agnostic to the number and the spacing of the disparity planes, we run the refinement module SegRefine independently for each disparity plane.
For continuous depth, whether full or selective, the distance between the planes is uniform, fixed and known before-hand.
Therefore, rather than applying a refinement to each binary segmentation, we apply a $3$D encoder-decoder module to the stacked outputs of the SegNet block, dubbed RegNet in Figure~\ref{fig:diagram}, (right).
Using a $3$D module RegNet leads to better results since it can use information from the neighboring disparity planes.
RegNet is closely inspired by the regularization module of GC-Net~\cite{kendall2017end}, with minor modifications described in the supplementary, also takes the left-image features as input for regularization.
The final step is the refinement proposed in StereoNet~\cite{khamis2018stereonet}, which takes the left image and the output of the AUC layer to generate the final disparity map, $M$. 

We first train Bi3DNet on the SceneFlow dataset.
We form a batch with $64$ stereo pairs and, for each of them, we randomly select a disparity plane for the segmentation.
We use a binary cross-entropy (BCE) loss and train for $1$k epochs.
To train for continuous depth estimation, we initialize FeatNet and SegNet with the weights trained for binary depth.
We form a batch with $8$ stereo pairs and use two different losses:
a BCE loss on the estimated segmentation confidence volume and SmoothL1 loss on the estimated disparity with weights of $0.1$ and $0.9$, respectively.
We train this network for $100$ epochs.
For the KITTI dataset, we fine-tune both networks starting from the weights for SceneFlow.
For both binary and continuous depth we use a batch size of $8$ and the procedure described above, but we train them for $5$k and $500$ epochs, respectively.
We use random crops of size $384 \times 576$, the Adam optimizer, and the maximum disparity of $192$ for all the trainings.

\section{Evaluation and Results}\label{sec:results}
%!TEX root = bi3d_cvpr20.tex

\begin{table}
    \centering
    \scalebox{0.92}{
    \begin{tabular}{cc|ccc}
        & & GA-Net & DeepPrunerFast &\multicolumn{1}{l}{Ours (Time)}\\
        \hline
        \parbox[t]{2mm}{\multirow{4}{*}{\rotatebox[origin=c]{90}{Levels}}} & 2 & 0.9654 & 0.9677& \multicolumn{1}{l}{0.9702 (5.3 ms)} \\
        &4 & 0.9302 & 0.9350 & \multicolumn{1}{l}{0.9372 (9.8 ms)}\\
        &8 & 0.8774 & 0.8826 &  \multicolumn{1}{l}{0.8909 (18.5 ms)}\\
        &16 & 0.8061 & 0.8066 & \multicolumn{1}{l}{0.8307 (36 ms)}
    \end{tabular}
    }
    \caption{Mean IOU for different depth quantizations (higher is better). For GA-Net and DeepPruner we quantize the full depth, see text. Our method is on par in terms of quality, but offers the ability to trade depth accuracy for latency. For reference DeepPrunerFast, which is faster than GA-Net, runs in $62$ ms.}\label{tab:quantization_iou}
\end{table}

\begin{table}[tb]
    \centering
    \scalebox{0.82}{
        \begin{tabular}{lccccccc}
            \hline
            &\multicolumn{3}{c}{Noc (\%)} &\multicolumn{3}{c}{All (\%)}\\
            Methods & \emph{bg} &\emph{fg} &\emph{all} &\emph{bg} &\emph{fg} &\emph{all} \\
            \hline
            MAD-Net~\cite{Tonioni2019} & 3.45 & 8.41 & 4.27 & 3.75 & 9.2 & 4.66\\ 
            Content-CNN~\cite{Luo2016} & 3.32 & 7.44 & 4.00 & 3.73 & 8.58 & 4.54 \\
            DipsNetC~\cite{mayer2016large} & 4.11 & 3.72 & 4.05 & 4.32 & 4.41 & 4.34 \\
            MC-CNN~\cite{zbontar2015computing}  & 2.48 & 7.64 & 3.33 & 2.89 & 8.88 & 3.89 \\
            GC-Net~\cite{kendall2017end}  & 2.02 & 3.12 & 2.45 & 2.21 & 6.16 & 2.87 \\
            CRL~\cite{Pang2017} & 2.32 & 3.68 & 2.36 & 2.48 & 3.59 & 2.67 \\ 
            PDS-Net~\cite{Tulyakov2018} & 2.09 & 3.68 & 2.36 & 2.29 & 4.05 & 2.58 \\
            PSM-Net~\cite{chang2018pyramid} & 1.71 & 4.31 & 2.14 &1.86 &4.62 &2.32 \\
            SegStereo~\cite{yang2018segstereo} & 1.76 & 3.70 & 2.08 & 1.88& 4.07 & 2.25 \\
AutoDispNet-CSS~\cite{saika2019_autodispnet} & 1.80 & 2.98 & 2.00 & 1.94 & 3.37 & 2.18 \\
EdgeStereo~\cite{song2018edgestereo} & 1.72 & 3.41 & 2.00 & 1.87 & 3.61 & 2.16 \\
DeepPruner-Best~\cite{duggal2019deeppruner} & 1.71 & 3.18 & 1.95 & 1.87 & 3.56 & 2.15 \\
GwcNet~\cite{Guo_2019_GwcNet}             & 1.61 & 3.49 & 1.92 & 1.74 & 3.93 & 2.11 \\
EMCUA~\cite{nie2019multi} & 1.50 & 3.88 & 1.90 & 1.66 & 4.27 & 2.09 \\
GA-Net-15~\cite{zhang2019ga} & 1.40 & 3.37 & 1.73 & 1.55 & 3.82 & 1.93 \\ 
CSPN~\cite{cheng2018depth} & 1.40 & 2.67 & 1.61 & 1.51 & 2.88 & 1.74 \\
\hline
\textbf{Ours} & \textbf{1.79} & \textbf{3.11} & \textbf{2.01} & \textbf{1.95} & \textbf{3.48} & \textbf{2.21}\\
\hline
\end{tabular}
}
\caption{Numerical results on the KITTI dataset. Our method outperforms several authoritative methods, such as PDS-Net~\cite{Tulyakov2018} and GC-Net~\cite{kendall2017end}.}
\label{tab:results}
%\vspace{-3mm}
\end{table}

Our goal is to offer a trade-off between latency and depth estimation accuracy---from binary depth, all the way to full continuous depth.
While no existing method is designed to deal with quantized depth,  we compare against two state-of-the-art (standard) stereo methods: 
GA-Net~\cite{zhang2019ga}, top-performing method on KITTI at the time of submission, and DeepPruner~\cite{duggal2019deeppruner}, because its ``fast'' version is among the fastest high-performing methods on KITTI.
Because these methods are not designed to estimate depth quantization directly, we run them to compute the full depth and then quantize it appropriately.
Because quantized depth estimation is a segmentation problem, we evaluate the results with the mean intersection over union (mIOU) with the depth labels used as classes.
Table~\ref{tab:quantization_iou} shows the results using $1$k randomly selected images from the Scene Flow Dataset~\cite{mayer2016large}.
Notice that our method yields a higher mIOU despite having access to less information (fewer disparity planes).
Perhaps more importantly, however, our method can run as fast as $5.3$ ms ($>180$ fps) for binary depth, or $9.8$ ms ($>100$ fps) with four levels of quantization (measured on a NVIDIA Tesla V100 with TensorRT).
The runtime depends on the specific implementation and hardware, and thus a direct comparison is not entirely fair. 
However, for reference, DeepPrunerFast reports $62$ ms.\footnote{The authors of GA-Net provide only the execution time for their fast version ($50$-$66$ ms for images with $42\%$ fewer pixels), but not the corresponding trained model.}
Figure~\ref{fig:coarse_depth} shows results of quantized depth for $4$, $8$, and $16$ levels of quantization.
Note that $4$ and $8$ levels are often enough to form a rough idea of the scene.

We also compare our algorithm for full, continuous depth estimation against state-of-the-art methods.
Tables~\ref{tab:resultssf}~and~\ref{tab:results} show results on Scene Flow and KITTI 2015 respectively.
Although the strength and true motivation for our method is its flexibility to the choice of the disparity range, our numbers are surprisingly close to state-of-the-art methods that specifically target these benchmarks.
On the Scene Flow dataset our EPE is the second best, Table~\ref{tab:resultssf}. 
Moreover, a visual comparison with recent GA-Net~\cite{zhang2019ga} and GwcNet~\cite{Guo_2019_GwcNet} shows a comparable quality, see Figure~\ref{fig:full_range_results}.

\section{Conclusions}\label{sec:conclusions}
%!TEX root = bi3d_cvpr20.tex

In this paper, we introduce a novel framework for stereo-based depth estimation.
We show that we can learn to classify which regions of the scene are in front or beyond a virtual fronto-parallel plane.
By performing many such tests and finding at which plane the classification of each pixel switches labels, we can can estimate accurate depth.
More importantly, however, it allows to effectively focus on specific depth ranges.
This can reduce the computational load or improve depth quality under a budget on the number of disparities that can be tested.
Although our focus is a flexible depth range, we also show that our method close or on par with recent, highly-specialized methods.

\section*{Acknowledgments}
The authors thank Zhiding Yu for the continuous discussions and Shoaib Ahmed Siddiqui for help with distributed training.
UCSB acknowledges partial support from NSF grant IIS 16-19376 and an NVIDIA fellowship for A. Badki.

\newpage
{\small
\bibliographystyle{ieee_fullname}
\bibliography{bi3d_cvpr20} 

\begin{thebibliography}{10}\itemsep=-1pt

\bibitem{chang2018pyramid}
Jia-Ren Chang and Yong-Sheng Chen.
\newblock Pyramid stereo matching network.
\newblock In {\em Proceedings of the IEEE Conference on Computer Vision and
  Pattern Recognition}, 2018.

\bibitem{cheng2018depth}
Xinjing Cheng, Peng Wang, and Ruigang Yang.
\newblock Depth estimation via affinity learned with convolutional spatial
  propagation network.
\newblock In {\em Proceedings of the European Conference on Computer Vision},
  2018.

\bibitem{collins1996space}
Robert~T. Collins.
\newblock A space-sweep approach to true multi-image matching.
\newblock In {\em Proceedings of the IEEE Conference on Computer Vision and
  Pattern Recognition}, 1996.

\bibitem{duggal2019deeppruner}
Shivam Duggal, Shenlong Wang, Wei-Chiu Ma, Rui Hu, and Raquel Urtasun.
\newblock {DeepPruner}: {L}earning efficient stereo matching via differentiable
  {P}atch{M}atch.
\newblock In {\em Proceedings of the IEEE Conference on Computer Vision and
  Pattern Recognition}, 2019.

\bibitem{geiger2010efficient}
Andreas Geiger, Martin Roser, and Raquel Urtasun.
\newblock Efficient large-scale stereo matching.
\newblock In {\em Proceedings of the Asian Conference on Computer Vision},
  2010.

\bibitem{Guo_2019_GwcNet}
Xiaoyang Guo, Kai Yang, Wukui Yang, Xiaogang Wang, and Hongsheng Li.
\newblock Group-wise correlation stereo network.
\newblock In {\em Proceedings of the IEEE Conference on Computer Vision and
  Pattern Recognition}, 2019.

\bibitem{hirschmuller2005accurate}
Heiko Hirschmuller.
\newblock Accurate and efficient stereo processing by semi-global matching and
  mutual information.
\newblock In {\em Proceedings of the IEEE Conference on Computer Vision and
  Pattern Recognition}, 2005.

\bibitem{hirschmuller2007evaluation}
Heiko Hirschmuller and Daniel Scharstein.
\newblock Evaluation of cost functions for stereo matching.
\newblock In {\em Proceedings of the IEEE Conference on Computer Vision and
  Pattern Recognition}, 2007.

\bibitem{im2019dpsnet}
Sunghoon Im, Hae-Gon Jeon, Stephen Lin, and In~So Kweon.
\newblock {DPSNet}: {E}nd-to-end deep plane sweep stereo.
\newblock {\em International Conference on Learning Representations}, 2019.

\bibitem{kendall2017end}
Alex Kendall, Hayk Martirosyan, Saumitro Dasgupta, Peter Henry, Ryan Kennedy,
  Abraham Bachrach, and Adam Bry.
\newblock End-to-end learning of geometry and context for deep stereo
  regression.
\newblock In {\em Proceedings of the IEEE Conference on Computer Vision and
  Pattern Recognition}, 2017.

\bibitem{khamis2018stereonet}
Sameh Khamis, Sean Fanello, Christoph Rhemann, Adarsh Kowdle, Julien Valentin,
  and Shahram Izadi.
\newblock Stereo{N}et: {G}uided hierarchical refinement for real-time
  edge-aware depth prediction.
\newblock In {\em Proceedings of the European Conference on Computer Vision},
  2018.

\bibitem{kim2013scene}
Changil Kim, Henning Zimmer, Yael Pritch, Alexander Sorkine-Hornung, and Markus
  Gross.
\newblock Scene reconstruction from high spatio-angular resolution light
  fields.
\newblock {\em ACM Transactions on Graphics (SIGGRAPH)}, 2013.

\bibitem{klaus2006segment}
Andreas Klaus, Mario Sormann, and Konrad Karner.
\newblock Segment-based stereo matching using belief propagation and a
  self-adapting dissimilarity measure.
\newblock In {\em IEEE International Conference on Pattern Recognition (ICPR)},
  2006.

\bibitem{kolmogorov2001computing}
Vladimir Kolmogorov and Ramin Zabih.
\newblock Computing visual correspondence with occlusions using graph cuts.
\newblock In {\em Proceedings of the IEEE International Conference on Computer
  Vision}, 2001.

\bibitem{Luo2016}
Wenjie Luo, Alexander~G. Schwing, and Raquel Urtasun.
\newblock Efficient deep learning for stereo matching.
\newblock In {\em Proceedings of the IEEE Conference on Computer Vision and
  Pattern Recognition}, 2016.

\bibitem{mayer2016large}
Nikolaus Mayer, Eddy Ilg, Philip Hausser, Philipp Fischer, Daniel Cremers,
  Alexey Dosovitskiy, and Thomas Brox.
\newblock A large dataset to train convolutional networks for disparity,
  optical flow, and scene flow estimation.
\newblock In {\em Proceedings of the IEEE Conference on Computer Vision and
  Pattern Recognition}, 2016.

\bibitem{Menze2015kittidataset}
Moritz Menze and Andreas Geiger.
\newblock Object scene flow for autonomous vehicles.
\newblock In {\em Proceedings of the IEEE Conference on Computer Vision and
  Pattern Recognition}, 2015.

\bibitem{nie2019multi}
Guang-Yu Nie, Ming-Ming Cheng, Yun Liu, Zhengfa Liang, Deng-Ping Fan, Yue Liu,
  and Yongtian Wang.
\newblock Multi-level context ultra-aggregation for stereo matching.
\newblock In {\em Proceedings of the IEEE Conference on Computer Vision and
  Pattern Recognition}, 2019.

\bibitem{Pang2017}
Jiahao Pang, Wenxiu Sun, Jimmy~SJ. Ren, Chengxi Yang, and Qiong Yan.
\newblock Cascade residual learning: A two-stage convolutional neural network
  for stereo matching.
\newblock In {\em ICCV Workshop on Geometry Meets Deep Learning}, 2017.

\bibitem{saika2019_autodispnet}
Tonmoy Saikia, Yassine Marrakchi, Arber Zela, Frank Hutter, and Thomas Brox.
\newblock {AutoDispNet}: {I}mproving disparity estimation with automl.
\newblock In {\em Proceedings of the IEEE International Conference on Computer
  Vision}, 2019.

\bibitem{scharstein2002taxonomy}
Daniel Scharstein and Richard Szeliski.
\newblock A taxonomy and evaluation of dense two-frame stereo correspondence
  algorithms.
\newblock {\em International Journal of Computer Vision}, 2002.

\bibitem{song2018edgestereo}
Xiao Song, Xu Zhao, Hanwen Hu, and Liangji Fang.
\newblock {EdgeStereo}: {A} context integrated residual pyramid network for
  stereo matching.
\newblock {\em Proceedings of the Asian Conference on Computer Vision}, 2018.

\bibitem{Tonioni2019}
Alessio Tonioni, Fabio Tosi, Matteo Poggi, Stefano Mattoccia, and Luigi
  Di~Stefano.
\newblock Real-time self-adaptive deep stereo.
\newblock In {\em Proceedings of the IEEE Conference on Computer Vision and
  Pattern Recognition}, 2019.

\bibitem{Tulyakov2018}
Stepan Tulyakov, Anton Ivanov, and Fran{\c{c}}ois Fleuret.
\newblock Practical deep stereo {(PDS):} {T}oward applications-friendly deep
  stereo matching.
\newblock In {\em Advances in Neural Information Processing Systems}, 2018.

\bibitem{yang2018segstereo}
Guorun Yang, Hengshuang Zhao, Jianping Shi, Zhidong Deng, and Jiaya Jia.
\newblock {SegStereo}: {E}xploiting semantic information for disparity
  estimation.
\newblock In {\em Proceedings of the European Conference on Computer Vision},
  2018.

\bibitem{zabih1994non}
Ramin Zabih and John Woodfill.
\newblock Non-parametric local transforms for computing visual correspondence.
\newblock In {\em Proceedings of the European Conference on Computer Vision},
  1994.

\bibitem{zbontar2015computing}
Jure Zbontar and Yann LeCun.
\newblock Computing the stereo matching cost with a convolutional neural
  network.
\newblock In {\em Proceedings of the IEEE Conference on Computer Vision and
  Pattern Recognition}, 2015.

\bibitem{zhang2019ga}
Feihu Zhang, Victor Prisacariu, Ruigang Yang, and Philip~H.S. Torr.
\newblock {GA-Net}: {G}uided aggregation net for end-to-end stereo matching.
\newblock In {\em Proceedings of the IEEE Conference on Computer Vision and
  Pattern Recognition}, 2019.

\end{thebibliography}
}

\newpage
\clearpage
\twocolumn[
  \begin{@twocolumnfalse}
{
   \newpage
   \null
   \begin{center}
      {\Large \bf {Bi3D}: {S}tereo Depth Estimation via Binary Classifications \par ({S}upplementary)}
      {
      \large
      \lineskip .5em
      \begin{tabular}[t]{c}
          
      \end{tabular}
      \par
      }
      \vskip .5em
      \vspace*{0pt}
   \end{center}
}
  \end{@twocolumnfalse}
]

\setcounter{section}{0}
\setcounter{figure}{0}
\setcounter{table}{0}
\setcounter{footnote}{0}

\section{Additional Details on Training}\label{sec:training}
%!TEX root = bi3d_cvpr20_supplementary.tex

\paragraph{FeatNet.}
The input images for our feature extraction network are normalized using mean and standard deviation of $0.5$ for each color channel.
This network is based on the feature extraction network of PSM-Net~\cite{chang2018pyramid}.
We first apply a conv layer with a stride of $3$ to get downsampled features.
This is followed by two conv layers, each with a stride of 1. 
We use $3\times3$ kernels, a feature-size of 32, and a ReLU activation.
This is followed by two residual blocks as proposed in PSM-Net~\cite{chang2018pyramid} each with a dilation of $2$, a stride of $1$ and a feature-size of $32$. 
This is followed by the SPP module and the final fusion operation as explained in PSM-Net~\cite{chang2018pyramid} to generate a $32$-channel feature map for the input image. 
We do not use any batch normalization layers in our training. 

\paragraph{SegNet.}
SegNet architecture takes as input concatenated left-image features and warped right-image features and generates a binary segmentation confidence map.
SegNet is a 2D encoder-decoder with skip-connections. 
The basic block of the encoder is composed of a conv layer that downsamples the features with a stride of $2$ followed by another conv layer with a stride of $1$. 
We use $3\times 3$ kernels.
We repeat this block $5$ times in the encoder.
The feature-sizes for these blocks are $128$, $256$, $512$, $512$ and $512$ respectively.
The basic block of the decoder is composed of a deconv layer with $4\times4$ kernels and a stride of $2$, followed by a conv layer with $3\times3$ kernels and a stride of $1$. 
This block is repeated $5$ times to generate the output at the same resolution of the input. 
The feature-sizes for these blocks are $512$, $512$, $256$, $128$ and $64$ respectively.
For all our layers we use a LeakyReLU activation with a slope of $0.1$. We do not use a batch normalization layer in our network.
We have a final conv layer with $3\times3$ kernels and without any activation to generate the output of SegNet.
Applying sigmoid to this output generates the binary segmentation confidence map.

\paragraph{RegNet.}
SegNet is applied independently for each input plane to generate the corresponding binary segmentation maps when we apply sigmoid operation.
RegNet is a 3D encoder-decoder architecture with residual connections and is based on GC-Net~\cite{kendall2017end}. 
The outputs of the SegNet corresponding to all input planes are concatenated to form an input 3D volume.
RegNet refines this volume using input left images features from FeatNet as a guide. 
Note that this architecture does not take the warped right image features. 
We first pre-process the input 3D volume using a conv layer with $3 \times 3\times 3$ kernels. 
We use a feature-size of $16$, a stride of $1$, and a ReLU activation for this step.
Then we concatenate the left image features with the features of each confidence map to generate an input volume with a feature-size of $48$.
This serves as input to a 3D encoder-decoder architecture. 
This architecture is same as the one proposed in Section 3.3 in GC-Net~\cite{kendall2017end}.
However, we use only half the features as used in the original GC-Net~\cite{kendall2017end} architecture and don't use any batch normalization layers in our architecture.
The output of this network is a refined volume at the same resolution as the input volume to this network. 
Applying sigmoid operation gives us the refined binary segmentation confidence volume.

\paragraph{DispRefine.}
Our disparity refinement network uses the left-image as a guide to refine the disparity map computed using area under the curve operation on binary segmentation confidence volume. 
We use the network proposed in StereoNet~\cite{khamis2018stereonet} for this purpose. 
However, we don't use any batch normalization layers in our network.

\paragraph{SegRefine.}
Our SegRefine network refines the upsampled output of the SegNet using the left-image as a guide. 
We first apply three conv layers each with $3\times3$ kernels with a stride of $1$ on the input left-image. 
We use a feature-size of $16$ for these layers. 
We apply ReLU activation on the first two layers. 
The third layer does not have any activation and gives us left-image features at the resolution of input left-image.
Note that these features need to be computed only once per stereo pair.
We then concatenate these feature maps with the upsampled output of SegNet and apply a single conv layer with a $3\times3$ kernel, a feature-size of $8$ and a LeakyReLU activation with a slope of 0.1.
A final conv layer with a $3\times3$ kernel and no activation generates the output of SegRefine network.
Applying sigmoid operation to this output gives us the binary segmentation confidence map at the resolution of input image.

\section{Supplementary Video}\label{sec:adadepth}
%!TEX root = bi3d_cvpr20_supplementary.tex

We explain the adaptive depth estimation application in the supplementary video, where we also further discuss the use of the area under the curve (AUC) formulation for disparity regression. 
The video is at \url{https://github.com/NVlabs/Bi3D}.

\end{document}